# A Few-Shot Metric Learning Method with Dual-Channel Attention for Cross-Modal Same-Neuron Identification

Wenwei Li[1], Liyi Cai[1], Wu Chen[1], Anan Li[1,2,3,*]

1 Britton Chance Center for Biomedical Photonics, Wuhan National Laboratory for Optoelectronics, MoE Key Laboratory for Biomedical Photonics, Huazhong University of Science and Technology, Wuhan 430074, China.

2 HUST-Suzhou Institute for Brainsmatics, JITRI, Suzhou 215123, China.

3 Key Laboratory of Biomedical Engineering of Hainan Province, School of Biomedical Engineering, Hainan University, Haikou 570228, China.

*Correspondence: aali@hust.edu.cn (A.L.)

## Abstract

In neuroscience research, achieving single-neuron matching across different imaging modalities is critical for understanding the relationship between neuronal structure and function. However, modality gaps and limited annotations present significant challenges. We propose a few-shot metric learning method with a dual-channel attention mechanism and a pretrained vision transformer to enable robust cross-modal neuron identification. The local and global channels extract soma morphology and fiber context, respectively, and a gating mechanism fuses their outputs. To enhance the model's fine-grained discrimination capability, we introduce a hard sample mining strategy based on the MultiSimilarityMiner algorithm, along with the Circle Loss function. Experiments on two-photon and fMOST datasets demonstrate superior Top-K accuracy and recall compared to existing methods. Ablation studies and t-SNE visualizations validate the effectiveness of each module. The method also achieves a favorable trade-off between accuracy and training efficiency under different fine-tuning strategies. These results suggest that the proposed approach offers a promising technical solution for accurate single-cell level matching and multimodal neuroimaging integration.

# 1. Introduction

In neuroscience research, achieving cross-modal matching and integration at the single-neuron level is of critical importance for understanding the intrinsic relationship between neuronal function and structure. By integrating data at the single-cell level, functional and structural imaging can be combined to comprehensively characterize both the physiological activity and morphological features of individual neurons, thereby providing powerful support for elucidating the structure-function organization of neural circuits.

Currently, cross-modal neuron matching methods largely rely on manual intervention. For example, Wang *et al.* proposed the 2-SPARSE method, which successfully reconstructed the full-brain morphology of sound-responsive neurons in the auditory cortex of awake mice [1]. Li *et al.* developed a strategy that integrates in vivo microscopic imaging and high-resolution fluorescence micro-optical sectioning tomography (fMOST) to map the whole-brain projection patterns of functionally characterized neurons in the somatosensory cortex, hippocampus, and substantia nigra pars compacta [2]. Zhou *et al.* proposed the use of vascular structures as auxiliary landmarks to aid in the registration of functional and structural imaging data[3]. More recently, in 2024, Li *et al.* introduced an automatic neuron matching method based on a graph model for aligning two-photon and fMOST imaging data[4]. However, this approach relies solely on the spatial distribution of neuron populations, which proves insufficient when faced with neurons that are spatially similar but morphologically distinct.

In contrast, neuronal morphological features serve as a key basis for similarity assessment and play a critical role in cross-modal data fusion at the single-cell level. Morphological feature recognition is essentially a pattern recognition problem, similar to face or fingerprint recognition. To address this challenge, this study introduces a neuron similarity discrimination network based on metric learning. By effectively mapping neuron data from different imaging modalities into a shared feature space, this approach ensures that features of the same neuron cluster closely, while those of different neurons remain distant, thus enabling accurate cross-modal neuron identification.

Nevertheless, due to the limited availability of annotated cross-modal neuron data, training a complex model from scratch often leads to overfitting. In recent years, the rapid development of deep learning—particularly Transformer-based architectures has driven the rise of transfer learning using pretrained models. In computer vision, several large-scale pretrained models such as Vision Transformer (ViT)[5], Swin Transformer[6, 7], Masked Autoencoder (MAE) [8], CLIP[9], Segment Anything (SAM) [10, 11], and DINO[12, 13] have demonstrated remarkable feature extraction capabilities and achieved state-of-the-art performance in tasks such as classification[14], object detection[15, 16], and image segmentation[17]. These models exhibit strong transferability and have been successfully adapted to the medical imaging domain. For instance, Zhao *et al.* discussed the application of CLIP in medical image analysis[18], while Zhu *et al.* built Medical SAM2 for medical image segmentation based on SAM2[19]. By leveraging the general features learned from large-scale datasets, these methods can be fine-tuned on small datasets to achieve significant performance gains with limited labeled data.

Metric learning is a fundamental technique in machine learning that aims to learn a feature embedding in which samples from the same class are close together while those from different classes are far apart. It has found widespread application in image recognition, retrieval, and classification, particularly demonstrating strong generalization and effectiveness in low-data scenarios.

Recent studies have explored the integration of metric learning with large pretrained models. For example, Aleksandr *et al.* proposed Hyp-DINO and Hyp-ViT, which utilize ViT and DINO as visual backbones and project features into hyperbolic space for improved similarity learning[20]. An *et al.* introduced the Unicom model, which combines clustering and feature compression techniques within a metric learning framework using the pretrained CLIP model. They performed clustering on the large-scale LAION-400M dataset to generate pseudo labels and used them to optimize representation learning. Their method outperformed existing approaches across multiple benchmarks, and the pretrained model was released publicly[21]. In this work, we adopt the Unicom model as the feature extractor and fine-tune it to perform neuron similarity classification with improved accuracy.

To this end, we propose a few-shot metric learning framework with dual-channel attention, built upon a pretrained vision model. This framework enhances sensitivity to cross-modal discrepancies and improves the robustness of neuron matching. Experimental results show that our method significantly improves cross-modal neuron identification accuracy under small-sample conditions, offering a novel and practical solution for high-resolution neuron-level data fusion.

## 2. Materials and methods

**2.1 Dataset Construction for Neuron Similarity Learning**

To build a robust metric learning model for neuron similarity, we utilized manually annotated neuron pairs as the ground truth for both training and evaluation. A total of 273 cross-modal neuron pairs were collected and annotated, where each pair comprises corresponding neuron regions imaged via two-photon and fMOST modalities, respectively. These were treated as positive sample pairs. Each paired sample image has a resolution of 150 × 150 pixels, corresponding to approximately 120 μm × 120 μm in real-world spatial scale, allowing coverage of the neuron soma as well as surrounding dendritic and axonal structures. Representative examples of positive pairs used for similarity learning are shown in Figure 1.

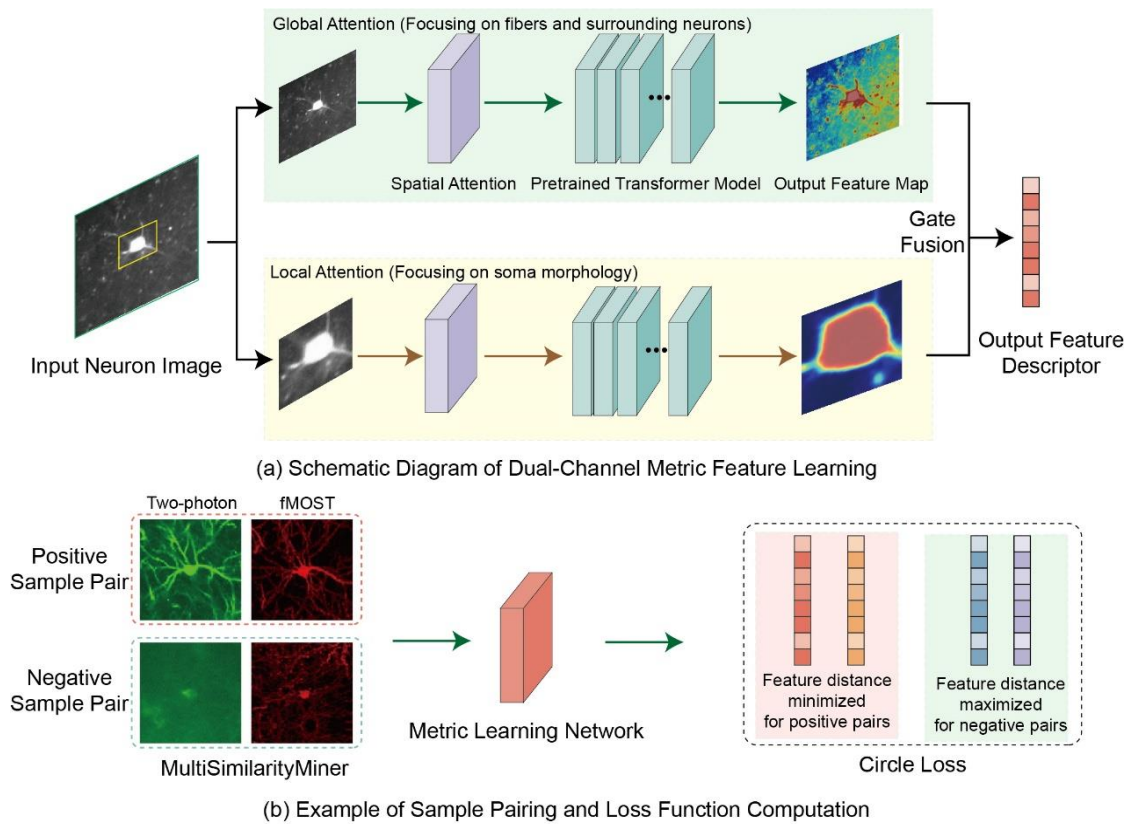

Figure 1. Network architecture and training strategy for metric learning.

For dataset partitioning, 220 pairs were allocated for model training and validation, among which 190 pairs were used for training and 30 pairs for validation. The remaining 53 pairs were reserved as a test set to evaluate the model's matching performance and generalization ability on unseen data.

**2.2 Dual-Channel Attention Mechanism for Integrating Global and Local Features**

Neuronal morphological similarity primarily depends on the shape of the soma and the patterns of surrounding fiber projections. To comprehensively extract multi-scale structural features from neuron images, we designed a dual-channel attention-based feature extraction network, as illustrated in Figure 1. This architecture integrates a local attention channel and a global attention channel, which respectively focus on soma morphology and spatial distribution, thereby enhancing the descriptive power and robustness in cross-modal neuron matching tasks.

Specifically, the local attention channel concentrates on the soma region, emphasizing contour shapes and fine-grained textures, which are crucial for capturing geometric and local structural characteristics of the neuron. In contrast, the global attention channel is designed to capture fiber trajectories and their spatial relationships with surrounding structures. This is achieved by incorporating a spatial attention module along with a pretrained Transformer model, enabling the network to model and capture broader contextual information. Each channel independently produces feature maps with different semantic granularities.

To effectively integrate these complementary features, we further introduce a gating-based fusion mechanism that adaptively weights and combines the outputs from both channels. The Gate module adjusts the fusion weights based on the importance of each feature stream, resulting in a unified high-dimensional feature representation optimized for similarity measurement.

**2.3 Network Training with Hard Sample Mining**

**(1) Cross-Modal Hard Sample Mining Strategy**

To enhance the model's ability to discriminate fine-grained features across imaging modalities, we introduce a cross-modal hard sample mining mechanism. During training, negative sample pairs that are challenging to distinguish are dynamically selected using the MultiSimilarityMiner strategy[22]. Let the sample set be denoted as $X = \{x_i\}$, then for a given sample xix_ixi, the corresponding hard negative sample set is defined as:

$$N_i = \{x_j | dist\left(f(x_i), f(x_j)\right) < \alpha, y_i \neq y_j\} \quad (1)$$

where $dist$ denotes the distance function in the feature space, $f(x_i)$ represents the output feature vector from the network, $y_i$ is the class label of sample $x_i$ and $\alpha$ is a distance threshold used for selecting hard samples.

This dynamic sampling strategy encourages the network to focus more on challenging examples during training, thereby improving its ability to distinguish subtle differences between neurons, especially under cross-modal conditions.

(2) Loss Function Design

To explicitly optimize both intra-class compactness and inter-class separability, we adopt the Circle Loss[23] as the objective function, which is defined as:

$$L_{circle} = log\left[1 + \sum_{i \in P}\sum_{j \in N} \exp\left(\gamma(s_n^j - s_p^i + m)\right)\right] \quad (2)$$

Here, $P$ and $N$ denote the sets of positive and negative sample pairs, respectively; $s_p^i$ is the similarity score for the $i$-th positive pair, $s_n^j$ is the similarity score for the $j$-th negative pair, $m$ is a margin parameter controlling the decision boundary, and $\gamma$ is a scale factor that adjusts the model's sensitivity to hard samples.

By explicitly weighting gradients based on sample proximity to the classification boundary, Circle Loss emphasizes difficult samples that are closer to the margin. This leads to tighter clustering of similar neurons and greater separation between dissimilar ones in the feature space, thereby significantly enhancing the model's fine-grained discriminative power.

## 3. Results

### 3.1 Evaluation of Method Effectiveness

During the testing phase of the metric learning model, the evaluation samples include both matched neuron pairs—manually verified across two-photon and fMOST modalities—and unmatched pairs. As the number of negative (unmatched) samples significantly exceeds that of positive samples, directly including all unmatched pairs as negatives would result in severe class imbalance, potentially biasing model evaluation. To address this, we designed a negative sample selection strategy to better assess the model's discriminative capability.

Specifically, the positive set comprises all 53 manually confirmed matched neuron pairs. The negative set is constructed at four times the size of the positive set, totaling 212 unmatched pairs, and is divided into two categories:

- Hard negatives: 106 unmatched pairs are selected from non-matching neuron combinations that exhibit the highest predicted similarity scores (i.e., most likely to confuse the model), serving as challenging distractors.

- Random negatives: Another 106 unmatched pairs are randomly sampled from two-photon and fMOST neuron datasets, aimed at increasing diversity and generalizability of the evaluation set.

This hybrid strategy ensures the test set covers both high-similarity distractors and general-case negatives, providing a more comprehensive reflection of the model's performance in realistic neuron matching scenarios.

During evaluation, the model's output—feature distances in the learned embedding space—is used to classify pairs as positive or negative. A threshold is applied: if the distance between a sample pair is below the threshold, it is classified as a positive (matched) pair; otherwise, it is classified as a negative (unmatched) pair. The final results are summarized in Table 1.

The model achieves a recall rate of 77.4%, ensuring a high proportion of true matches are correctly identified, while maintaining a specificity (rejection rate) of 90.1%, indicating strong ability to correctly reject non-matching pairs. These results demonstrate that the proposed method performs well in distinguishing neuron correspondence under a balanced test design.

**Table 1. Confusion Matrix for Metric Learning Evaluation**

|  | True Positive (Matched) | True Negative (Unmatched) |
|---|---|---|
| Predicted as Positive | 41 | 21 |
| Predicted as Negative | 12 | 191 |

### 3.2 Neuron Retrieval Based on Metric Learning

We evaluated the proposed method on both two-photon calcium imaging and fMOST datasets by performing a neuron retrieval task. Specifically, neurons imaged via two-photon microscopy were used as query samples, and the most similar neuron samples were retrieved from the fMOST neuron database based on distances in the learned feature space. Figure 2 illustrates representative retrieval results for six neuron samples. The left column shows the query neurons from two-photon imaging, the middle column shows the top retrievals from fMOST, and the right column displays the corresponding ground-truth labels.

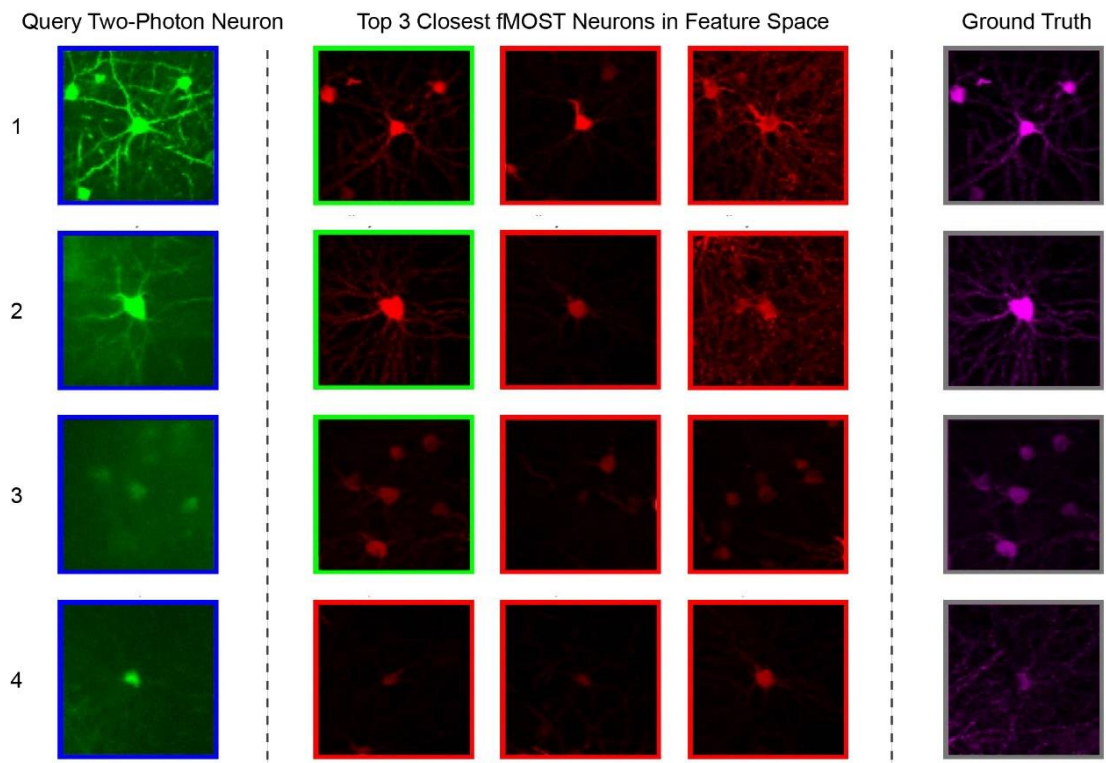

**Figure 2. Neuron retrieval results based on metric learning.**

In Figure 2, samples 1 through 3 represent successful retrieval cases where the retrieved neurons from the feature space closely match the ground truth. In sample 1, the query neuron exhibits clear structure with well-preserved surrounding neuronal features. The metric learning model accurately retrieves the most similar fMOST neuron, and the remaining candidates differ significantly in feature distance. Sample 2 lacks contextual background and presents weak fiber details, but the soma boundary is prominent enough to support correct retrieval. In sample 3, although the query neuron has blurry soma boundaries and indistinct fibers, the spatial arrangement of surrounding neurons provides auxiliary cues that contribute to successful matching.

In contrast, sample 4 is a failure case. The query neuron appears extremely blurred, with no discernible soma or fiber features, nor any surrounding neuronal context. As a result, the model fails to extract meaningful representations and cannot retrieve the correct match.

These observations demonstrate that the proposed metric learning framework is capable of learning to represent soma shapes, fiber structures, and contextual neuron distributions—akin to the way humans visually recognize neuronal identity. Thus, the method effectively extracts comprehensive neuron image features and retrieves accurate cross-modal correspondences.

### 3.3 Ablation Studies

To thoroughly evaluate the contribution of each module in the proposed framework to the neuron matching task, we conducted a series of ablation experiments, focusing on the dual-channel feature extractor, Gate fusion module, Circle Loss function, and hard sample mining strategy.

**Table 2. Ablation Study for Neuron Similarity Metric Learning**

| Method | Configuration | Objective |
| --- | --- | --- |
| Proposed Method | Dual-channel + Gate Fusion + Circle Loss + MultiSimilarityMiner | Full model (ours) |
| Standard Sampling | All positive/negative pairs, random sampling | Evaluate effectiveness of hard sample mining |
| Triplet Loss | Replacing Circle Loss with Triplet Loss | Evaluate loss function effectiveness |
| Local Branch Only | Global channel removed | Evaluate contribution of local branch |
| Global Branch Only | Local channel removed | Evaluate contribution of global branch |

First, we examined the impact of the sampling strategy by comparing our method—based on MultiSimilarityMiner—with a conventional random sampling baseline, which pairs positive and negative samples without considering sample difficulty. In contrast, MultiSimilarityMiner dynamically selects hard negatives, helping the model focus on more challenging training cases.

Second, we compared Circle Loss with the traditional Triplet Loss, a common metric learning objective based on triplets of anchor, positive, and negative samples. Circle Loss was shown to provide better class compactness and inter-class separation, improving overall discriminative power.

Third, we assessed the contribution of the dual-branch architecture by removing either the local or global attention branch and training the model with only one channel. This analysis helps isolate the effect of each component on model performance.

The results of the ablation study are visually presented using Top-k accuracy curves in Figure 4, where the curve reflects the proportion of correctly matched neurons among the top-k retrieval results. An ideal retrieval model achieves high accuracy at a small k value, with a rapidly rising curve.

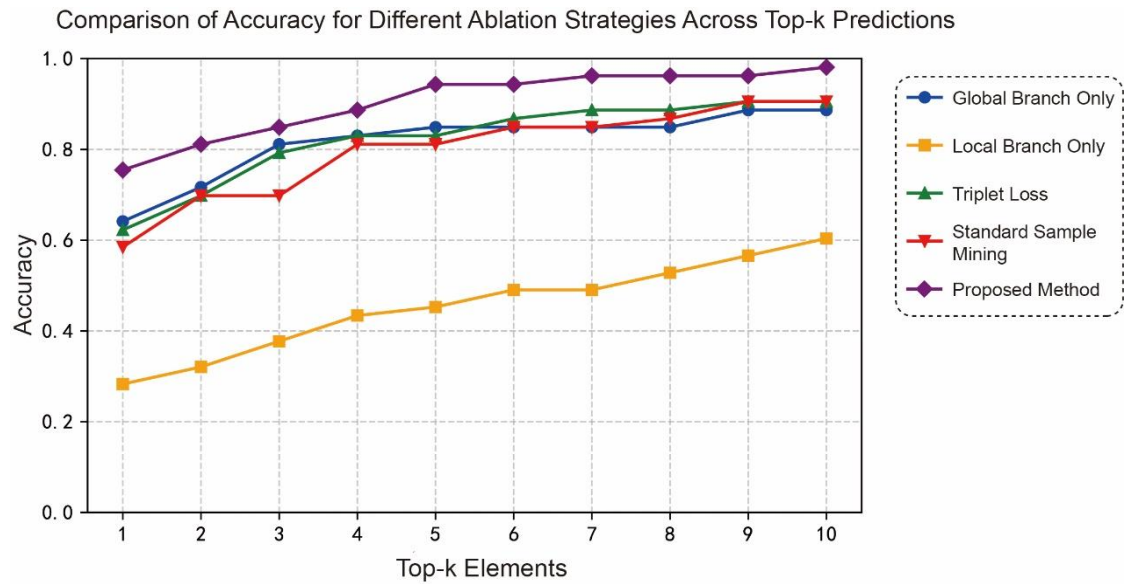

Figure 4. Top-k accuracy comparison for different model variants.

From the results, we observe that: The local-only model underperforms due to lack of global structural context; The global-only model lacks fine-grained detail needed for soma-level matching; Models using standard sampling or Triplet Loss fail to focus on hard-to-distinguish samples, leading to inferior accuracy; The full model, integrating dual-channel features, Gate fusion, Circle Loss, and hard sample mining, achieves the best performance, significantly outperforming all ablation variants.

**3.4 Comparison of Fine-Tuning Strategies**

Since there is currently a lack of established benchmarks for neuron similarity metric learning, this study leverages pretrained models from the field of computer vision. To explore the impact of different fine-tuning strategies on task performance, we design and evaluate four representative methods: linear fine-tuning, partial parameter fine-tuning, LoRA fine-tuning, and full parameter fine-tuning. Together with the original pretrained model and our proposed model, a total of six configurations are compared. Table 3-8 summarizes the setup of each method.

**Table 3-8. Comparison of Methods for Neuron Similarity Learning**

| Method | Parameter Setup | Description |
| --- | --- | --- |
| Original Model | Pretrained weights, no fine-tuning | Baseline model |
| Linear Fine-Tuning | Fine-tune only the final linear layer | Lightweight tuning strategy |
| Partial Fine-Tuning | Fine-tune the last four Transformer blocks | Evaluate impact of partial adaptation |
| LoRA Fine-Tuning | Low-rank adaptation with r = 4, 8, 16 | Efficient fine-tuning with reduced parameter count |
| Full Fine-Tuning | Fine-tune all model parameters | Traditional full transfer learning |
| Proposed Method | Full framework developed in this study | Dual-channel + Gate + Circle Loss + Hard mining |

**Linear Fine-Tuning**: In this approach, the feature extraction backbone of the pretrained model is kept fixed, and only a linear layer is added at the end to serve as a task-specific classifier or regressor. The core idea is to leverage the rich representations learned during pretraining and adapt to downstream tasks by training only a small number of parameters. This strategy offers low computational cost and training stability, especially in scenarios with limited data or high similarity between the pretraining and downstream tasks.

**Partial Parameter Fine-Tuning**: This strategy utilizes the hierarchical structure of the pretrained Transformer. The feature extractor used in this study contains 8 self-attention modules. To retain deep representations while enabling task-specific adaptation, we freeze the first 4 self-attention blocks and fine-tune only the last 4 self-attention blocks along with the final linear layer.

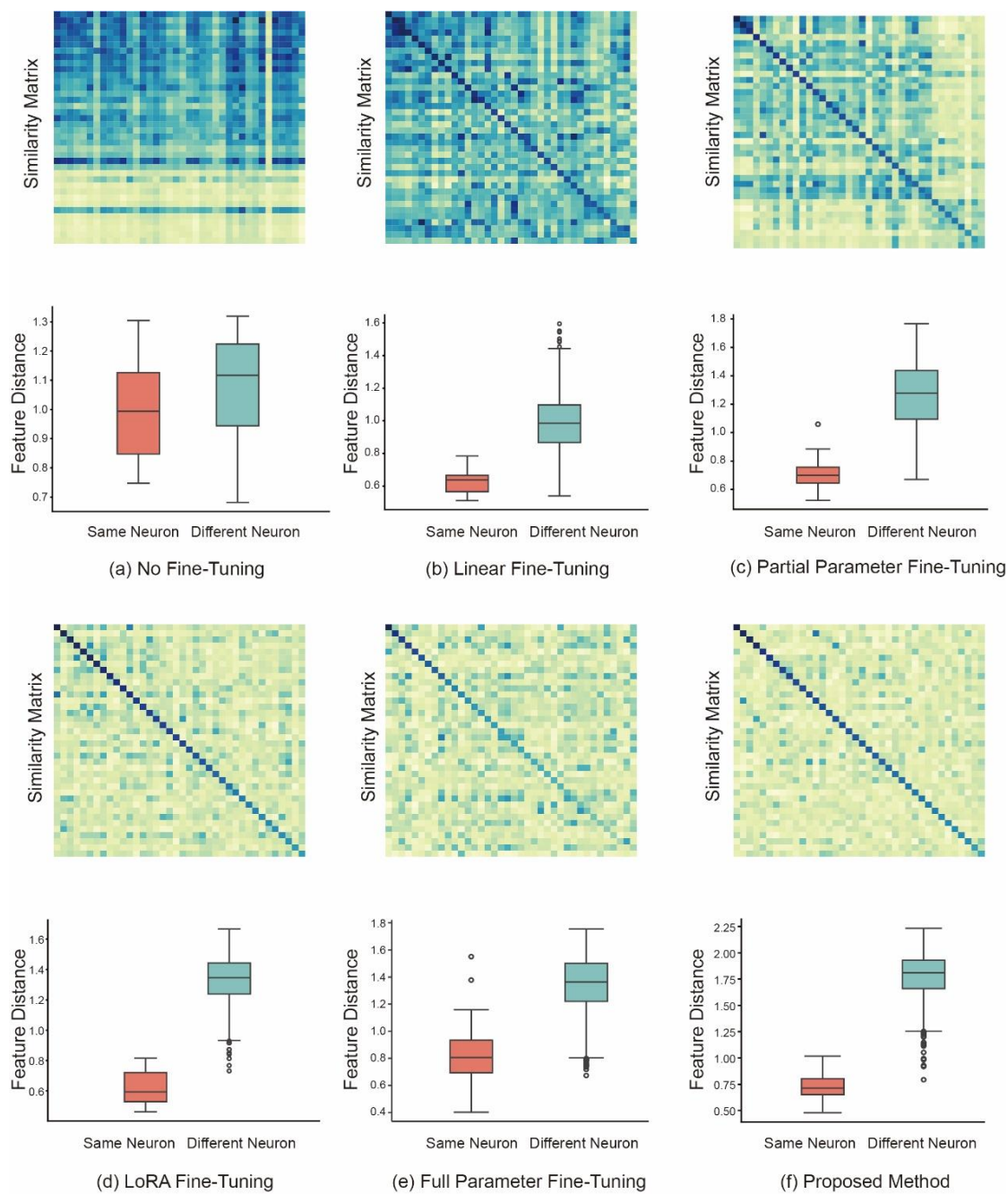

**LoRA Fine-Tuning**: LoRA (Low-Rank Adaptation)[24] is a parameter-efficient fine-tuning technique widely adopted in natural language processing and computer vision. The main idea is to preserve the pretrained weights and inject low-rank matrices into linear transformations to enable adaptable learning. In this study, all original linear layers are frozen, and only the newly introduced low-rank matrices are updated during fine-tuning. To investigate the impact of rank constraints, we test r = 4, 8, and 16. LoRA balances performance with resource efficiency by significantly reducing the number of trainable parameters while injecting task-specific adaptability.

**Full Parameter Fine-Tuning**: This traditional transfer learning strategy fine-tunes all parameters of the pretrained model, enabling full model flexibility for downstream adaptation. While this method allows the network to learn task-specific patterns extensively, it also risks overfitting, especially when training data is limited. We use full fine-tuning as a comparative baseline to evaluate the effectiveness and applicability of other fine-tuning strategies across different scenarios.

To objectively assess the performance of each method, Top-k accuracy curves are used for systematic comparison, as shown in Figure 5.

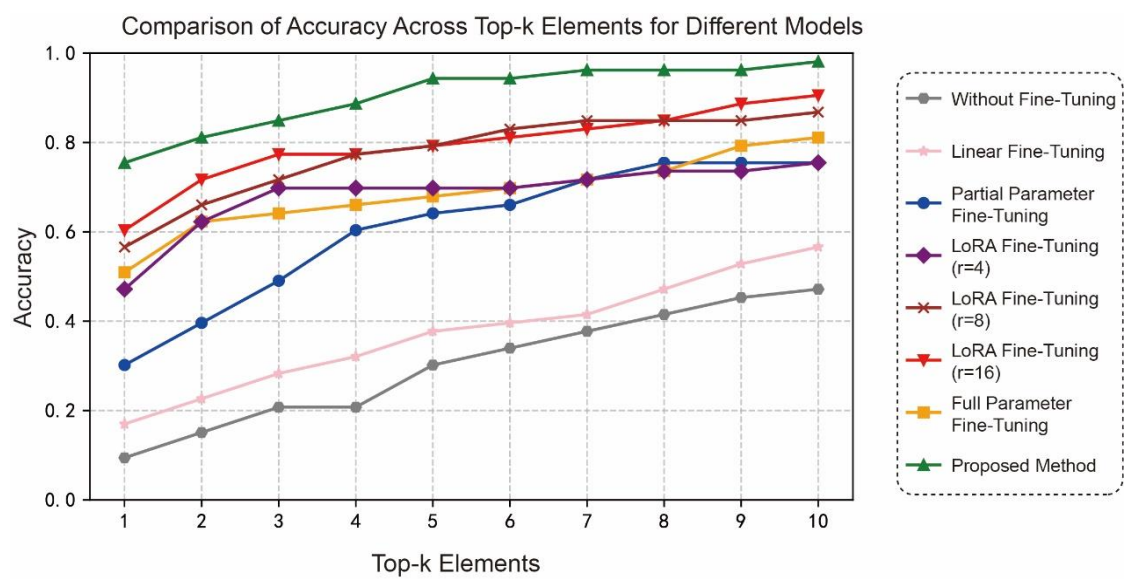

**Figure 5. Effects of Fine-Tuning Strategies on Neuron Similarity Accuracy.**

The results show that our proposed method consistently outperforms all baseline fine-tuning strategies across all Top-k evaluation metrics, demonstrating its superior capability in learning cross-modal feature similarity for neurons.

Figure 6 further visualizes the feature similarity heatmaps and box plots of similarity distributions for different strategies. Each method is evaluated in terms of distinguishing between same-neuron pairs and different-neuron pairs, with LoRA results shown for r = 16. In the heatmaps (top row), diagonal elements represent matched two-photon and fMOST neurons, while off-diagonal elements correspond to mismatched pairs. A greater contrast between diagonal and off-diagonal brightness indicates higher discriminative power. In the box plots (bottom row), less overlap between the distributions of positive and negative samples implies better feature separability.

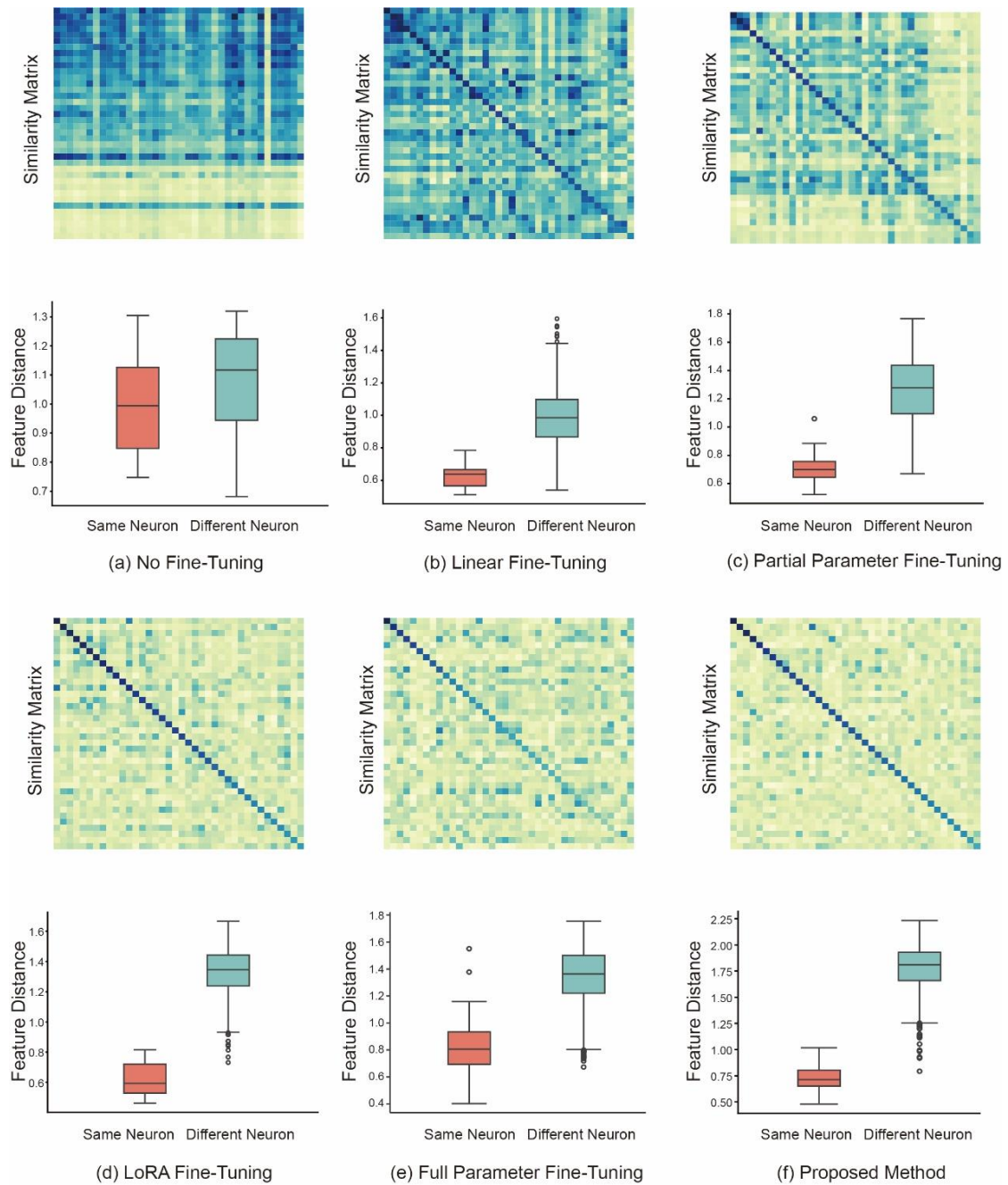

Figure 6. Evaluation of Fine-Tuning Strategies: Similarity Heatmaps and Feature Distributions.

From the visualization, it is evident that The original pretrained model fails to distinguish matched from unmatched pairs; Linear fine-tuning lacks sufficient adaptation due to minimal parameter updates; Full fine-tuning may lead to overfitting due to the large number of trainable parameters; Partial fine-tuning and LoRA significantly improve results by updating a subset of critical parameters; The proposed method, which integrates both local and global feature representations, achieves the best performance in differentiating between matched and unmatched neuron pairs.

**3.5 Network Visualization**

**(1) Visualization of Network Attention**

To intuitively illustrate the regions of input images that the model focuses on when determining neuron similarity, we employed Grad-CAM (Gradient-weighted Class Activation Mapping)[25] for visual analysis of the deep neural network. Grad-CAM is a gradient-based visualization technique that generates class activation heatmaps aligned with the spatial dimensions of the input image, thereby highlighting the critical regions that contribute to the model's decision.

The implementation proceeds as follows: a target convolutional module within the model is selected, and forward and backward hook functions are registered to capture the corresponding feature maps and gradients. During the forward pass, the input image is propagated through the network to obtain the embedded features while recording the outputs of the selected module. During the backward pass, gradients are computed with respect to a scalar objective defined on the output features, and the spatially averaged gradients are used as channel-wise weights. These weights are then linearly combined with the corresponding feature maps to form the initial class activation map. ReLU is applied to remove negative activations, and the result is resized to match the input image dimensions and normalized to the [0,1] range.

To enhance interpretability, the normalized heatmap is rendered as a pseudocolor image and blended with the original input using linear fusion. By adjusting the transparency parameter α, one can clearly observe which regions the model emphasizes during the matching task. This visualization approach not only helps validate whether the model's attention is focused on neuronal structures, but also offers insights into model behavior, providing an intuitive reference for future design and refinement.

As shown in Figure 7, the local attention branch predominantly focuses on edge contours and soma boundaries, while the global attention branch captures fiber branches and their spatial relationships with surrounding structures. These results confirm that the proposed dual-channel attention structure effectively captures and fuses multi-scale features of neurons, thereby significantly improving cross-modal matching performance.

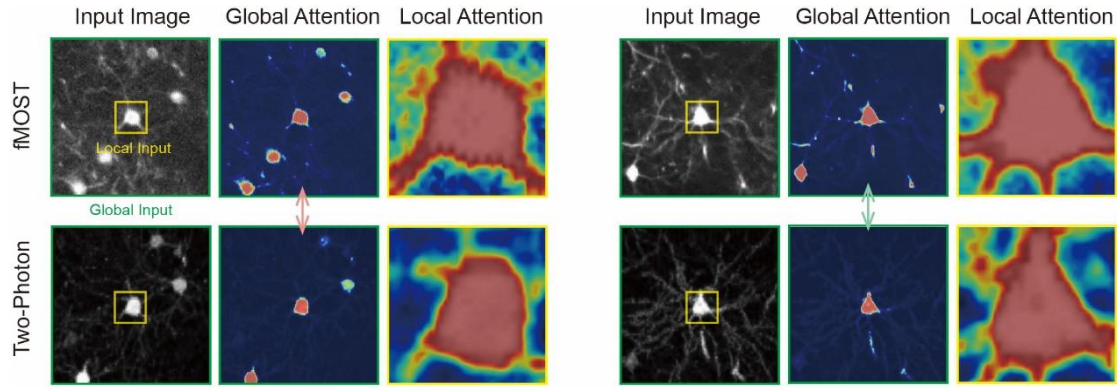

Figure 7. Attention Visualization of the Dual-Channel Network

**(2) Visualization of Feature Space via Dimensionality Reduction**

To further assess the effectiveness of the proposed method in aligning cross-modal neuron features, we applied the t-distributed Stochastic Neighbor Embedding (t-SNE)[26] algorithm to perform nonlinear dimensionality reduction on the high-dimensional neuron features extracted from two-photon and fMOST imaging modalities. By projecting these features into a 2D space, we can intuitively observe the relative distribution of matched pairs across modalities.

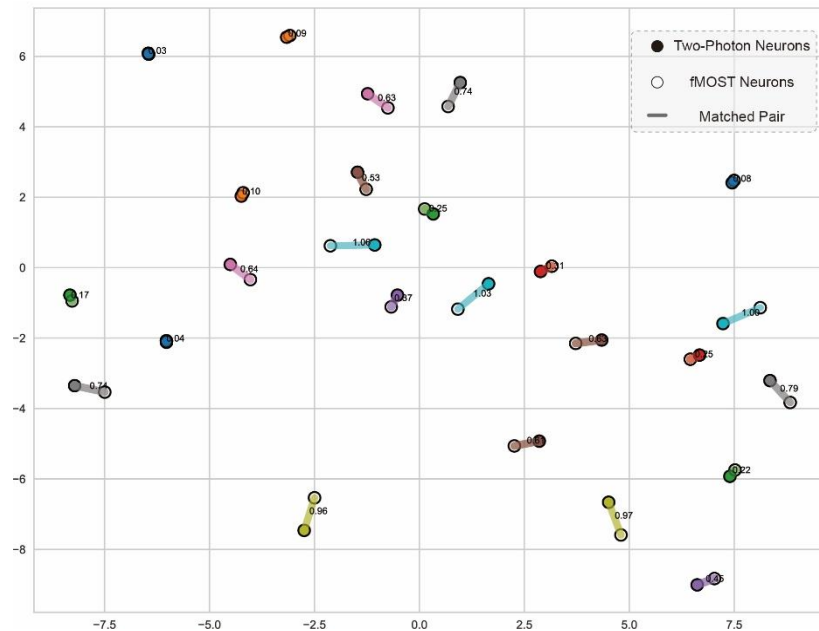

Figure 8. Visualization of Feature Embedding for Paired Neurons After Dimensionality Reduction

Specifically, we first extract the deep features of each neuron sample and label them according to their source modality and pairing index. We then apply t-SNE to reduce all samples to a common 2D embedding space and visualize several selected neuron pairs. In Figure 8, filled circles represent two-photon neurons and hollow circles represent fMOST neurons. Points of the same color correspond to a successfully matched neuron pair, and lines connect each pair.

From the visualization, it is evident that the matched neuron pairs exhibit good spatial aggregation in the t-SNE space, with most successfully matched pairs located close to one another in the low-dimensional embedding. This result verifies the consistency and accuracy of our method in achieving cross-modal feature alignment within the learned representation space.

**3.6 Comparative Evaluation of Model Performance**

To quantitatively assess the effectiveness of the proposed model, we compared our method against several conventional and deep learning-based similarity measures. The evaluated methods include: Normalized Mutual Information (NMI), Cosine Similarity, Structural Similarity Index (SSIM), Pearson Correlation Coefficient, and Mean Squared Error (MSE).

Figure 9 presents the neuron similarity results obtained using different similarity metrics, including the proposed metric learning-based approach, MSE, NMI, Pearson correlation, cosine similarity, and SSIM. The evaluation consists of similarity matrices and kernel density estimation (KDE) curves, providing a comprehensive comparison of the methods in quantifying neuron similarity across two-photon and fMOST imaging modalities.

In the similarity matrices, diagonal elements represent matched neurons (i.e., same neuron across modalities), and off-diagonal elements represent mismatched neuron pairs. In the KDE plots, the blue curve represents the distribution of similarities for matched neurons, while the orange curve represents mismatched neurons.

The upper section of Figure 9 shows the similarity matrices generated by each method. Diagonal elements indicate similarity scores between paired neurons across two-photon and fMOST imaging, while off-diagonal elements indicate scores between non-matching pairs. A strong contrast between diagonal and off-diagonal values suggests good discriminative performance.

The lower section shows KDE curves for each method. The degree of separation and overlap between blue (positive pairs) and orange (negative pairs) distributions reflects how well each method distinguishes matched from unmatched neurons.

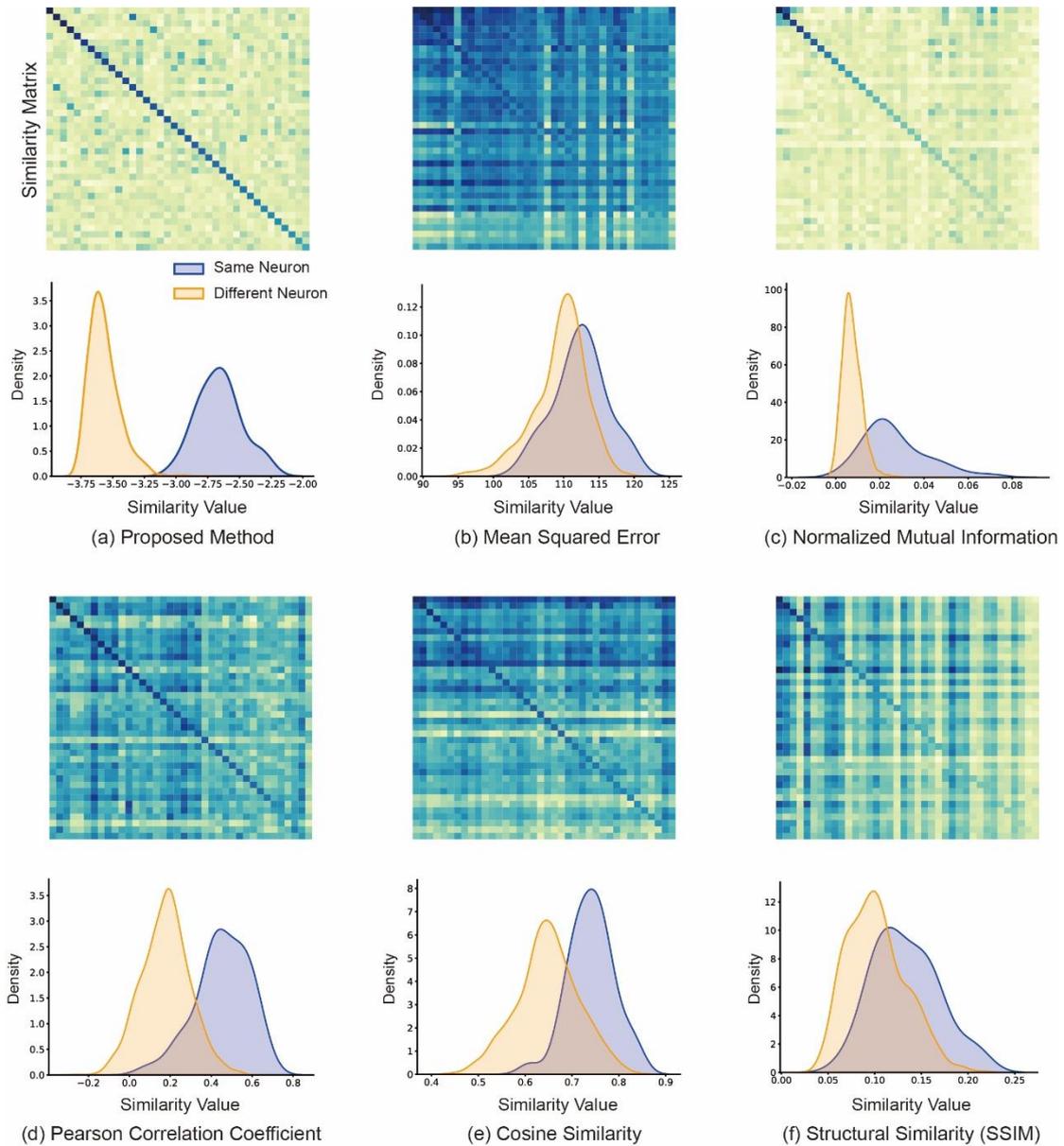

Figure 9. Neuron similarity matrices and KDE curves using different methods.

Figure 9(a) corresponds to the proposed method, which integrates global and local features via a dual-channel attention mechanism. The similarity matrix shows distinctively higher diagonal values, indicating that the model effectively differentiates same-neuron pairs. The KDE curves exhibit minimal overlap between positive and negative distributions, confirming the strong discriminative power of the metric learning framework.

Figure 9(c) shows the results of Normalized Mutual Information (NMI). While the similarity matrix demonstrates reasonable distinction in some high-similarity positive pairs, it fails to separate low-similarity positives from negatives. The KDE plot reveals moderate overlap, indicating NMI has limited ability to distinguish low-similarity matches, though it performs relatively well for dissimilar pairs.

Figure 9(d) and 9(e) correspond to Pearson correlation and cosine similarity, respectively. Their similarity matrices show some ability to separate diagonal and off-diagonal elements. However, KDE curves for both methods indicate significant overlap between high-similarity negatives and low-similarity positives. This suggests that while these metrics offer basic separation, they are inadequate for handling complex, cross-modal neuron data.

Figure 9(b) and 9(f) correspond to Mean Squared Error (MSE) and Structural Similarity Index (SSIM). In both cases, the similarity matrices show no clear diagonal dominance, and the KDE plots exhibit substantial overlap between blue and orange curves. These results indicate that MSE and SSIM perform poorly, failing to capture structural differences and similarities between neuron pairs across modalities.

In summary, the proposed metric learning method outperforms all baseline methods in distinguishing between same and different neurons. While minor overlap still exists in the KDE curves, the separation between positive and negative pairs is significantly more pronounced. These results validate the method's effectiveness and provide a solid foundation for accurate neuron similarity assessment in cross-modal imaging scenarios.

## Discussion

In this study, we developed a metric learning-based framework for neuron similarity prediction, which demonstrates strong potential for application in cross-modal neuron matching tasks. Notably, the proposed method achieves robust and accurate performance even under conditions of limited labeled data and small sample sizes. This approach not only provides a novel solution for neuron matching between two-photon and fMOST imaging but also establishes a technical foundation for single-neuron level integration of cross-modal data.

Despite the promising results achieved in both methodological design and experimental validation, several aspects remain open for further exploration and refinement:

Feature Representation Adaptability: The current morphological feature extraction strategies exhibit limited adaptability across different imaging modalities. Future work could investigate unified representations that generalize well to diverse imaging techniques.

Attention Mechanism Interpretability: While the proposed dual-channel attention mechanism empirically enhances feature discriminability, the theoretical basis for its effectiveness remains underexplored. Deeper analysis of its underlying mechanism—possibly augmented with visualization techniques—could provide further insight and validation.

Dataset Limitations: The present experiments are conducted on a relatively small dataset. Extensive evaluations on larger and more diverse cross-modal neuron datasets are needed to thoroughly assess the generalizability and robustness of the proposed approach.

Architectural and Optimization Improvements: There remains room for improvement in both network architecture and loss function design. Incorporating more efficient computational modules or adaptive loss strategies may further enhance performance while reducing computational cost.

In conclusion, the proposed metric learning framework for cross-modal neuron similarity discrimination shows promising application potential in preliminary experiments. However, continued research is necessary to strengthen its theoretical underpinnings and facilitate deployment in real-world scenarios, particularly in neuroscientific research and clinical decision support systems.

## CRediT authorship contribution statement

Wenwei Li: Writing – review & editing, Validation, Software, Methodology, Conceptualization. Liyi Cai: Data curation. Wu Chen: Data curation. Anan Li: Writing – original draft, Writing – review & editing, Investigation, Formal analysis, Conceptualization, Supervision, Methodology, Funding acquisition.

## Acknowledgements

We would like to thank the MOST group of Britton Chance Center for Biomedical Photonics,

Wuhan National Laboratory for Optoelectronics, MoE Key Laboratory for Biomedical Photonics, Huazhong University of Science and Technology. This study was supported by STI 2030-Major Projects (2021ZD0201002), the National Natural Science Foundation of China Grants (T212015), and the National Natural Science Foundation of China Grants (82102235).

**Declaration of competing interest**

The authors declare no competing interests.

# Cite